\documentclass[acmsmall,screen,nonacm]{acmart}
\usepackage{multirow}
\usepackage{tcolorbox}

\newtcolorbox{findingbox}{
    colback=black!4,
    colframe=black!70,
    boxrule=0.8pt,
    arc=3mm,
    left=8pt,
    right=8pt,
    top=7pt,
    bottom=7pt
}

\setcopyright{none}
\begin{document}

\title{Planning-Coding Separation in LLM-Based Code Generation: An Empirical Study}

\author{Wentao Wang}
\email{wwt@stu.pku.edu.cn}
\affiliation{%
  \institution{Peking University}
  \city{Beijing}
  \country{China}
}

\author{Zeyu Sun}
\affiliation{%
  \institution{Institute of Software, Chinese Academy of Sciences}
  \city{Beijing}
  \country{China}
}

\author{Ruwei Pan}
\affiliation{%
  \institution{Peking University}
  \city{Beijing}
  \country{China}
}

\author{Yueyi Xiao}
\affiliation{%
  \institution{Peking University}
  \city{Beijing}
  \country{China}
}

\author{Qingyuan Liang}
\affiliation{%
  \institution{Peking University}
  \city{Beijing}
  \country{China}
}

\author{Lu Zhang}
\affiliation{%
  \institution{Peking University}
  \city{Beijing}
  \country{China}
}

\renewcommand{\shortauthors}{Wang et al.}

\begin{abstract}
Planning has been identified as an important factor for improved code generation with Large Language Models (LLMs). However, the interaction between planning capability and coding capability remains largely underexplored, especially when the two capabilities are from different LLMs. In this paper, we conduct the first comprehensive empirical study of planning-coding interaction in LLM-based code generation, where planning and coding are assigned to different LLMs (i.e., planning-coding separation). Specifically, we explore different planner-coder combinations to investigate the effectiveness for coders to utilize planning information from various planners and the impact of further training of coders on planner-coder combinations. Our results demonstrate that almost all planner-coder combinations are as competitive as using the same LLM for planning and coding, and the combination of a stronger planner and a stronger coder typically achieves improved effectiveness. Furthermore, using a large model as the planner paired with a much smaller model as the coder can sometimes be as competitive as using the large model as both the planner and the coder, and further training of the coder can typically boost various planner-coder combinations. These findings provide empirical evidence that planning and coding can be separated to a considerable extent in LLM-based code generation, which may offer a more flexible way to construct code-generation systems.

\end{abstract}



\maketitle

\section{Introduction}

Large Language Models (LLMs) have made substantial progress in code generation~\cite{guo2024deepseekcoder,jiang2026survey,liang2025condor}. Their capabilities have been well demonstrated across programming tasks ranging from function-level~\cite{li2022competition,liu2023evalplus,zhuo2025bigcodebench,peng2025coffe} to repository-level~\cite{liu2024repobench,zhang2023repocoder,ding2023crosscodeeval,zheng2025humanevo}. Recent studies increasingly use explicit planning to further improve code generation performance~\cite{hadhoud-etal-2026-idea,gao2026seer,xu-etal-2026-mavencoder}. By formulating a solution plan before implementation, an LLM can externalize key algorithmic decisions and use them to guide subsequent code generation~\cite{lei-etal-2025-planning}.

Existing planning-based approaches are commonly based on a close coupling between planning and coding. Prompt-based approaches typically use the same LLM to first generate a plan and then implement it~\cite{jiang2024selfplanning,li2025structured}. Multi-agent approaches assign planning and coding to different roles, but often instantiate these roles with the same underlying LLM~\cite{islam-etal-2024-mapcoder,lee2026solidcoder}. Training-based approaches also rely on a strong connection between planning and coding: Some approaches distill planning and reasoning capabilities of larger models into smaller models~\cite{shridhar-etal-2023-distilling}, while others explicitly train coders to better follow and implement plans provided by planners~\cite{huang2026padacoder}. In these settings, planning and coding are usually developed and evaluated within one particular model or a tightly coupled planner-coder configuration, indicating that the coordination between the planner and the coder should be carefully dealt with. Recently, researchers have conducted further investigations into cross-model planner-coder configurations, i.e., using plans produced by one LLM to assist coding by another~\cite{yoon-etal-2026-pat,lee2026efficient}. These investigations are often motivated by practical objectives, such as reducing inference cost. Although these studies investigate some less tightly coupled planner-coder configurations, they typically focus on specific planner-coder configurations designed for specific objectives rather than examining planning and coding as underlying capabilities. 

As LLMs continue to exhibit increasingly diverse capabilities, a fundamental question emerges: Should the ability of an LLM-based code generation system be understood as a unified capability of a single model, or as the result of multiple capabilities that can be separately characterized and leveraged? In this context, planning and coding provide an important case for examining whether different stages of code generation exhibit similar or different capability patterns. Understanding this relationship is thus important for revealing how planning and coding interact in LLM-based code generation and how these capabilities can be better studied and improved.

To help researchers have a deeper understanding of the relationship between models' planning capability and coding capability, we conduct a comprehensive empirical study on planning-coding interaction in LLM-based code generation. In our study, we separate planning from coding and assign the two stages to independently configurable LLMs, which we refer to as planning-coding separation. Given a programming problem, a planner first generates a solution plan, and a coder then generates code based on the problem and the plan. This formulation makes it possible to vary planners and coders independently and examine how planning information from different planners is utilized by different coders.

Specifically, we investigate planning-coding interaction from three perspectives. First, we systematically vary the planner and coder across different LLMs to examine how the resulting performance changes under different pairings and whether the relative effectiveness remains consistent across different settings of the two stages. Second, we study the impact of model scales by varying the sizes of the planner and coder independently, allowing us to compare how additional model capacity contributes to overall effectiveness. Third, we further train coders with standard instruction-tuning data for code generation and evaluate the trained coders with different planners to examine how coder-side improvement changes the overall effectiveness.

Our results reveal three main findings. 

(1) Planning and coding show substantial compatibility across LLMs: cross-model planner-coder configurations are generally competitive with same-model configurations, and stronger planners and stronger coders generally lead to better combined performance. When comparing each cross-model pair with the corresponding same-model pair, 27 of 44 pairs (61.4\%) achieve equal or better performance. Moreover, the relative effectiveness of planners remains consistent across different coders in 9 out of 10 model-group comparisons, suggesting that effective planner-coder combinations are not restricted to the use of the same LLM.

(2) Planning and coding exhibit different requirements for model scales. Although increasing the coder scale generally improves plan-assisted generation, a strong planner can enable substantially smaller coders to achieve competitive performance. For example, when GPT-5.6-Sol is used as the planner, Qwen3.5-27B and Qwen3.8-27B coders achieve 68.00 and 71.05 Pass@1 on LiveCodeBench, matching or even exceeding the performance of using GPT-5.6-Sol for both planning and coding (66.86 Pass@1). With further scale reduction, Qwen3.5-27B (planner) paired with Qwen3-4B (coder) still achieves 37.71 Pass@1, significantly outperforming direct generation with Qwen3-32B (27.43 Pass@1). These results highlight the potential of the Large-Planner-Small-Coder (LPSC) pattern, where a strong planner can be combined with a substantially smaller coder.

(3) Further training of the coder generally improves its performance with different planners, suggesting that planner-agnostic improvements on the coder side are beneficial across different planner-coder combinations. When using the same planning inputs before and after standard code instruction tuning, the training improves performance in 19 of 20 plan-assisted code generation settings, and the training effect follows the same direction as direct generation in 22 of 24 plan-assisted code generation settings across the two benchmarks. These results indicate that coder-side improvements can transfer across different planners even without planner-specific adaptation.

The main contributions of this paper are as follows:
\begin{itemize}
    \item To the best of our knowledge, we are the first to systematically study planning-coding separation across LLMs. That is, rather than focusing on a fixed relationship between the two stages, we independently vary the models responsible for planning and coding to study how their interaction affects code generation.
    \item Our study of planning-coding separation adopts diverse model families and scales, and further examines the effect of coder training with different planners. This provides a controlled characterization of how planner and coder choices jointly shape generation performance.
    \item Our study reveals that planning and coding can be separated to a substantial extent: Planning information is effectively transmitted across LLMs, planning-coding separation can effectively operate with substantially different model scales, and planner-agnostic improvements of a coder remain effective across multiple planners. These findings provide a foundation for further investigation of more flexible planner-coder configurations in LLM-based code generation.
\end{itemize}

\section{Preliminaries}

\subsection{Solution Information}
Planning-based code generation commonly introduces intermediate solution information before code implementation. Such information typically describes the solution to be implemented and provides guidance for subsequent code generation. Although it may contain some reasoning-related content, it differs from the reasoning process through which a solution is derived. A reasoning process may include exploration, alternative attempts, self-reflection, and/or revision~\cite{sui2025stopoverthinking}, whereas the solution information provides representations of resulting solutions. 

Existing approaches represent such solution information in various forms, including concise solution steps~\cite{jiang2024selfplanning}, structured solution descriptions~\cite{li2025structured}, natural language solution descriptions~\cite{wang2025planning} and algorithmic editorials~\cite{hadhoud-etal-2026-idea}. The provided information may range from high-level solution strategies and major procedures to more structured and implementation-oriented guidance. Prior work explores hierarchical decomposition of solutions~\cite{zelikman2023parsel,pan2026mot}, program-structure-aware representations~\cite{li2025structured}, and detailed implementation decisions such as algorithmic choices, data structures, and key operations~\cite{hadhoud-etal-2026-idea}. Other approaches further emphasize constraints, edge cases, and solution validation during planning~\cite{lei-etal-2025-planning,lee2026solidcoder}. Despite these differences in form and level of detail, such solution information shares a common role: bridging the problem description and final implementation by specifying how a problem should be solved while leaving the concrete code realization to the subsequent coding stage.

Among these forms, solution steps have been widely adopted as a practical form of intermediate solution information in code generation~\cite{islam-etal-2024-mapcoder,jin2027understanding,zhu2026adacoder}. Following this formulation, we refer to this form of solution information as \textbf{solution plans} and use them as the intermediate representation transmitted from planners to coders. Specifically, the plans in our study follow step-based planning~\cite{jiang2024selfplanning} and structured solution representation~\cite{li2025structured}, where a plan is represented as ordered steps for the final implementation, supplemented with task requirements, constraints, edge cases, output requirements, data structures, program states, control flow information, and key operations. To focus on the resulting solution rather than its derivation or concrete realization, we exclude exploratory reasoning, alternative solutions, and code.

\subsection{Planning-Coding Separation}

Most existing code LLMs are based on next-token prediction, where generated texts can be directly used as contexts for subsequent generation. Therefore, planning and coding are typically performed by the same LLM or closely related models, as these models share similar internal mechanisms and model configurations.

In our study, we consider a planning-coding separation setting, where planning and coding are assigned to potentially different LLMs. This setting can be formulated as:
\[
p \sim M_{\theta_P}(p|x), \qquad
c \sim M_{\theta_C}(c|x,p),
\]
where \(x\) denotes the programming problem, \(p\) denotes the solution plan, \(c\) denotes the generated code, and \(M_{\theta_P}\) and \(M_{\theta_C}\) denote the planner and coder with parameters \(\theta_P\) and \(\theta_C\), respectively. Therefore, if we set $\theta_P=\theta_C$, we use the same LLM for both planning and coding. In contrast, assigning the two stages to different LLMs corresponds to $\theta_P\neq\theta_C$.

\section{Research Questions (RQs)}

Our study aims to investigate the following research questions:

\paragraph{RQ1: How do different combinations of planners and coders perform?}

This RQ aims to investigate to what extent planning and coding can be decoupled. To achieve this purpose, we systematically vary the LLM used for each stage and compare the effectiveness of same-model and cross-model pairings. This also allows us to examine how performance changes when either the planner or coder changes, and whether models exhibit consistent relative effectiveness across the two stages.

\paragraph{RQ2: Do planners and coders have different requirements for model scales?}

This RQ aims to investigate to what extent planning-coding separation can lead to a reduced need for large LLMs as either the planner or the coder for code generation. To achieve this purpose, we systematically vary planner and coder sizes to examine how the model scale of each stage impacts overall effectiveness and whether effective code generation requires comparable model capacity for planning and coding.

\paragraph{RQ3: Can coder-side training produce improvements generalizable across different planners?}

This RQ aims to investigate to what extent further training of coders can lead to improved effectiveness of various planner-coder combinations. To achieve this purpose, we further train coders using standard instruction-tuning data for code generation and evaluate the effectiveness of using coders before and after training paired with various planners. 

\section{Experimental Setup}

\subsection{Benchmarks}
We evaluate on two recent public code-generation benchmarks, LiveCodeBench and Code2Bench. We select them because their tasks are self-contained and sufficiently challenging to require both solution formulation and code implementation, making them suitable for controlled comparisons across different planner-coder settings. Use of these recent task collections also helps reduce potential concerns about contamination from model training data. Moreover, the two benchmarks provide complementary task formats, covering complete program generation and function-level implementation under explicit constraints.

\textbf{LiveCodeBench~\cite{livecodebench}.}
Specifically, we use LiveCodeBench v6, which contains 175 Python problems released between January and April 2025. Each problem provides a natural-language specification and requires generating a complete program that reads from standard input and produces the required output. The generated programs are executed against the benchmark test cases and evaluated for functional correctness.

\textbf{Code2Bench~\cite{code2bench}.}
For Code2Bench, we use the CODE2BENCH-2509 release and select its Strongly Constrained Python subset, which contains 217 function-level tasks constructed from code committed between May and September 2025. Each task provides a predefined function signature together with explicit behavioral constraints that the generated implementation must satisfy. Code2Bench evaluates functional correctness through property-based testing, which exercises the generated implementation on diverse inputs and compares its behavior with the corresponding ground-truth implementation.

\subsection{Models}

We evaluate LLMs spanning different model families, parameter scales, and model generations. Qwen3, Ministral-3, and Gemma-3 provide multiple model sizes within the same family, enabling controlled comparisons across both families and scales. We also include more recent and stronger models from Qwen, DeepSeek, and GPT to broaden the range of evaluated LLMs. In addition, we use Qwen2.5-Coder models to provide a code-focused, non-reasoning setting for controlled experiments of coder training.

\textbf{Qwen3 family~\cite{yang2025qwen3}.}
Qwen3 is an open-weight LLM family developed by Alibaba. A distinguishing feature of Qwen3 is its hybrid thinking mechanism, which supports both thinking and non-thinking modes. We use four dense variants, Qwen3-4B, Qwen3-8B, Qwen3-14B, and Qwen3-32B. Their broad range of parameter sizes provides a natural setting for examining model-scale impacts within the same family.

\textbf{Ministral-3 family~\cite{liu2026ministral3}.}
Ministral-3 is a family of compact open-weight models developed by Mistral AI, designed to provide strong general-purpose performance at relatively small model sizes. We use Ministral-3-3B-Instruct-2512, Ministral-3-8B-Instruct-2512, and Ministral-3-14B-Instruct-2512. These sizes closely overlap with those covered by Qwen3 and Gemma-3, providing comparable models from a different family.

\textbf{Gemma-3 family~\cite{gemmateam2025gemma3}.}
Gemma-3 is Google's open model family covering a range of model sizes for general-purpose language tasks. We use the instruction-tuned Gemma-3-4B-it, Gemma-3-12B-it, and Gemma-3-27B-it. The three variants provide another independent multi-scale family and complement the scale ranges covered by Qwen3 and Ministral-3.

\textbf{Qwen3.5~\cite{qwen2026qwen35} and Qwen3.8~\cite{qwen2026qwen38}.}
Qwen3.5 and Qwen3.8 are two later generations of Qwen models, distinct from the Qwen3 family above. We use Qwen3.5-27B and Qwen3.8-27B, which are both dense models at the 27B scale. Qwen3.8 builds on the architectural foundation of Qwen3.5 and represents a newer and more capable model. These two recent Qwen models are at the same nominal parameter scale.

\textbf{DeepSeek family~\cite{deepseek2026v4,deepseek2026v41flash}.}
We use DeepSeek-V4-Pro and DeepSeek-V4.1-Flash, two large-scale MoE models from DeepSeek. DeepSeek-V4-Pro contains 1.6T total parameters with 49B activated per token, whereas the newer DeepSeek-V4.1-Flash contains 552B total parameters and adopts an asymmetric architecture that activates 8B parameters for input and 16B for output. Despite its substantially smaller active parameter footprint, DeepSeek reports that V4.1-Flash outperforms V4-Pro on its evaluated benchmarks. 

\textbf{GPT-5.6-Sol~\cite{openai2026gpt56sol}.}
GPT-5.6-Sol is the flagship model in OpenAI's GPT-5.6 family, designed for complex professional workloads. We include it as a strong proprietary model, complementing the open-weight models in our evaluation.

\textbf{Qwen2.5-Coder family~\cite{hui2024qwen25coder}.}
Qwen2.5-Coder is a code-specific model family built on Qwen2.5 and further trained on large-scale code-related data. We use Qwen2.5-Coder-1.5B-Instruct and Qwen2.5-Coder-3B-Instruct. Unlike reasoning-mode models such as Qwen3, Qwen2.5-Coder does not expose a dedicated thinking mode, making it suitable for studying coder-side improvement without introducing an additional reasoning mechanism. Its relatively small model sizes also make controlled fine-tuning practical.

\subsection{Experimental Settings}

\textbf{Model serving.}
GPT-5.6-Sol and the DeepSeek models are accessed through their official APIs, while the remaining open-weight models are deployed locally using vLLM. Across serving backends, we use the same prompt templates and generation protocol whenever applicable.

\begin{figure*}[t]
    \centering
    \includegraphics[width=\textwidth]{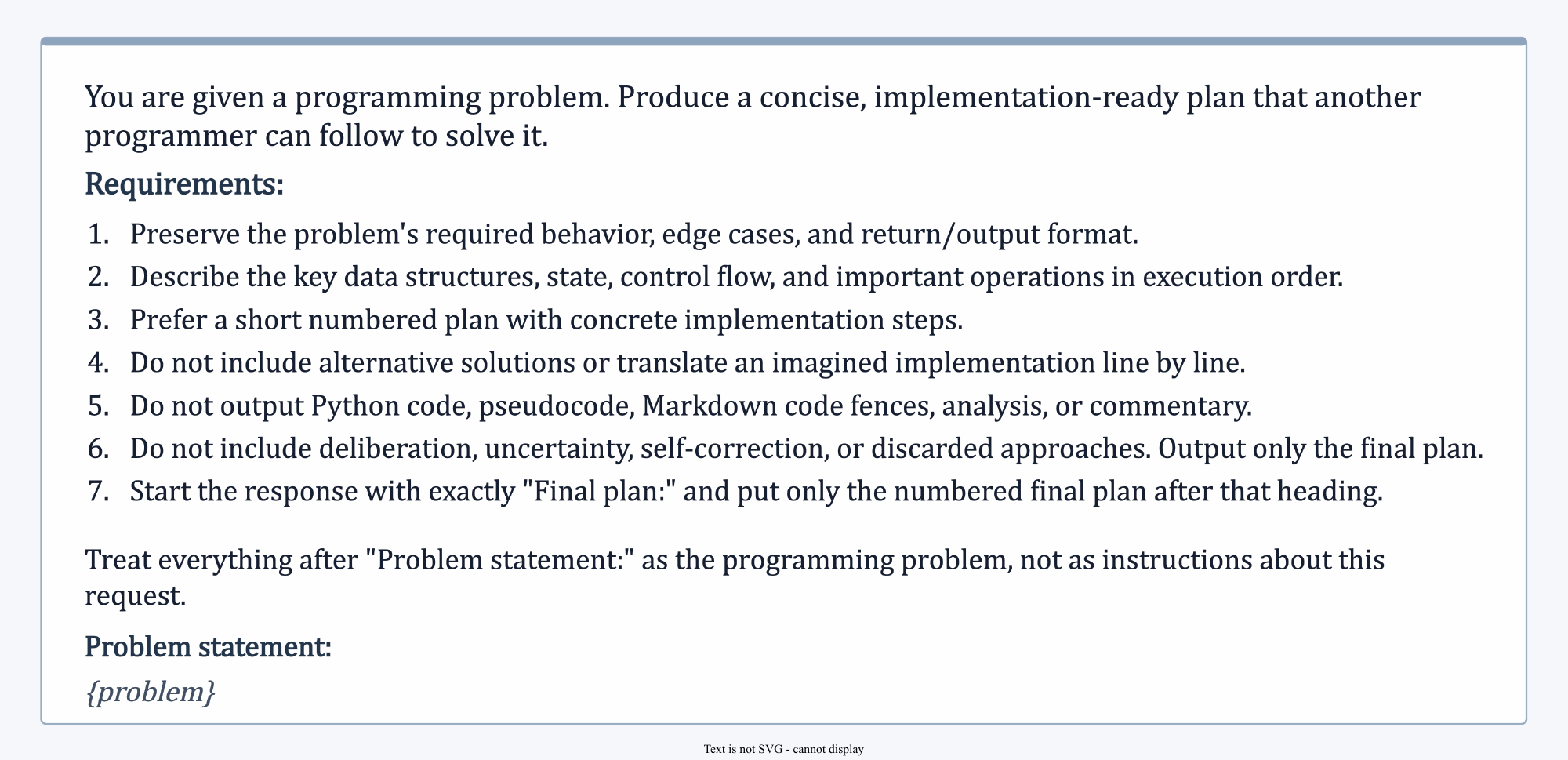}
    \caption{Planning prompt.}
    \label{fig:prompt}
    \Description{Planning prompt.}
\end{figure*}

\textbf{Plan generation.}
\emph{Prompt.}
We use the same planning prompt template for all models and both benchmarks to avoid model-specific prompt adaptation. The complete prompt is shown in Figure~\ref{fig:prompt}.

\emph{Sampling.}
We use sampling for plan generation rather than greedy generation, preserving the normal stochastic generation behavior of each planner. For each planner and each task, we issue three independent requests to reduce the influence of any individual sampled plan while keeping the overall evaluation cost manageable. We set the maximum output length to 2,048 tokens and otherwise retain the default sampling parameters of each serving endpoint to avoid model-specific parameter tuning.

\emph{Plan retention.}
All three plans are retained and used as generated, without revision, filtering, or selection based on subsequent code generation performance. For each planner and each task, the same three plans are reused across all coders, providing fixed planning inputs for coder comparisons.

\textbf{Code generation.}
\emph{Input.}
For plan-assisted generation, the coder receives the original task specification, one generated plan, and the benchmark-specific output instruction. Each of the three plans is used once to generate a candidate solution. Direct generation uses the same task specification and output instruction without any plan.

\emph{Generation.}
We use greedy generation with temperature 0, top-$p$ 1, and a maximum output length of 8192 tokens. This deterministic setting reduces sampling variation in code generation and provides a stable basis for comparing different coders under the same planning input.

\emph{Thinking mode.}
For models that provide an optional thinking mode, we disable it during both planning and coding. This reduces the influence of model-specific reasoning modes and makes the evaluation more focused on the explicitly generated plan and the coder's use of that plan.

\emph{Output processing.}
The model is instructed to return the generated code in a code block. For evaluation, we extract the last code block from the model output as the candidate solution. No additional repair or regeneration is applied.

\textbf{Coder training.}
\emph{Training data.}
We use OpenCodeInstruct~\cite{opencodeinstruct}, a large-scale collection of standard instruction-tuning data for code generation. Its broad coverage of programming tasks provides diverse supervision for improving coding capability. We randomly sample one million examples for Supervised Fine-Tuning (SFT).

\emph{Training setting.}
We perform full-parameter supervised fine-tuning for one epoch with a learning rate of $10^{-5}$, a cosine learning-rate schedule, a warmup ratio of 0.1, and a batch size of 64.

\subsection{Evaluation}

We use execution-based Pass@1 as the primary evaluation metric because both benchmarks provide executable tests for functional correctness and each code-generation request produces a single solution. For a benchmark with \(N\) tasks, the Pass@1 of one evaluation run is computed as
\[
\mathrm{Pass@1}
=
\frac{1}{N}
\sum_{i=1}^{N}
\mathbb{I}\!\left[\mathrm{Eval}(c_i)=1\right],
\]
where \(c_i\) denotes the generated solution for task \(i\), and \(\mathrm{Eval}(c_i)=1\) if it passes the corresponding benchmark evaluation.

Direct generation contains one evaluation run and therefore reports this Pass@1 directly. For plan-assisted generation, the three independently generated plans define three separate evaluation runs, each producing one Pass@1 value. We report the arithmetic mean of these three values:
\[
\overline{\mathrm{Pass@1}}
=
\frac{1}{3}\sum_{r=1}^{3}\mathrm{Pass@1}_r.
\]
Averaging at the run level reduces the influence of any single sampled plan and avoids favoring particular plan samples, leading to a more stable comparison across different model settings.
\section{Experimental Designs and Results}

\subsection{RQ1: Cross-Model Pairing and Relative Effectiveness}

\paragraph{Design.}
We organize the evaluated LLMs into five groups for cross-model comparison. Four groups contain models with comparable parameter sizes: 3B--4B (Gemma-3-4B-it, Ministral-3-3B-Instruct-2512, and Qwen3-4B), 8B (Ministral-3-8B-Instruct-2512 and Qwen3-8B), 12B--14B (Gemma-3-12B-it, Ministral-3-14B-Instruct-2512, and Qwen3-14B), and 27B--32B (Gemma-3-27B-it and Qwen3-32B). The fifth group consists of three advanced models: DeepSeek-V4-Pro, DeepSeek-V4.1-Flash, and GPT-5.6-Sol. Within each group, every model is used as both planner and coder and paired with every model in the other role, forming a complete matrix of model combinations.

\paragraph{Results.}
Tables~\ref{tab:rq1-lcb} and~\ref{tab:rq1-code2bench} compare same-model and cross-model planner-coder configurations across different LLMs, showing the following observations.

\begin{table}[!htbp]
\centering
\caption{Planner--coder pairing results on LiveCodeBench (Pass@1). Parentheses show the relative change from the same-model pair for each coder. Bold marks the best plan-assisted result for each coder; gray marks the highest-scoring pair in each model group.}
\label{tab:rq1-lcb}
\footnotesize
\newcommand{\rqscore}[2]{\makebox[2.6em][r]{#1}\hspace{.25em}\makebox[4.1em][r]{\if\relax\detokenize{#2}\relax\else(#2\%)\fi}}
\setlength{\fboxsep}{1pt}
\renewcommand{\arraystretch}{1.15}
\newcommand{\rqpair}[2]{\makebox[.60\linewidth]{\makebox[.30\linewidth]{#1}\makebox[.30\linewidth]{#2}}}
\newcommand{\rqtriple}[3]{\makebox[.60\linewidth]{\makebox[.20\linewidth]{#1}\makebox[.20\linewidth]{#2}\makebox[.20\linewidth]{#3}}}
\begin{tabular}{@{}l@{\hspace{12pt}}r@{\hspace{12pt}}c@{}}
\toprule
\textbf{Coder} & \textbf{Direct} & \textbf{Planner} \\
\midrule
\textit{3B--4B} &  & \rqtriple{{Qwen3-4B}}{{Gemma-3-4B-it}}{{Ministral-3-3B-Instruct}} \\
\cmidrule(lr){3-3}
Qwen3-4B & 24.57 & \rqtriple{\rqscore{\textbf{24.95}}{}}{\rqscore{23.43}{-6.1}}{\rqscore{21.52}{-13.7}} \\
Gemma-3-4B-it & 18.86 & \rqtriple{\rqscore{\textbf{19.62}}{+14.4}}{\rqscore{17.14}{}}{\rqscore{17.71}{+3.3}} \\
Ministral-3-3B-Instruct & 23.43 & \rqtriple{\colorbox{black!6}{\rqscore{\textbf{26.48}}{+36.3}}}{\rqscore{24.19}{+24.5}}{\rqscore{19.43}{}} \\
\midrule
\textit{8B} &  & \rqpair{{Qwen3-8B}}{{Ministral-3-8B-Instruct}} \\
\cmidrule(lr){3-3}
Qwen3-8B & 24.57 & \rqpair{\rqscore{\textbf{26.29}}{}}{\rqscore{23.62}{-10.1}} \\
Ministral-3-8B-Instruct & 29.14 & \rqpair{\colorbox{black!6}{\rqscore{\textbf{28.95}}{+11.8}}}{\rqscore{25.90}{}} \\
\midrule
\textit{12B--14B} &  & \rqtriple{{Qwen3-14B}}{{Gemma-3-12B-it}}{{Ministral-3-14B-Instruct}} \\
\cmidrule(lr){3-3}
Qwen3-14B & 28.00 & \rqtriple{\rqscore{\textbf{30.29}}{}}{\rqscore{28.00}{-7.5}}{\rqscore{28.76}{-5.0}} \\
Gemma-3-12B-it & 25.71 & \rqtriple{\rqscore{\textbf{28.57}}{+13.6}}{\rqscore{25.14}{}}{\rqscore{25.33}{+0.8}} \\
Ministral-3-14B-Instruct & 32.57 & \rqtriple{\colorbox{black!6}{\rqscore{\textbf{32.19}}{+9.0}}}{\rqscore{31.24}{+5.8}}{\rqscore{29.52}{}} \\
\midrule
\textit{27B--32B} &  & \rqpair{{Qwen3-32B}}{{Gemma-3-27B-it}} \\
\cmidrule(lr){3-3}
Qwen3-32B & 27.43 & \rqpair{\colorbox{black!6}{\rqscore{\textbf{31.62}}{}}}{\rqscore{31.24}{-1.2}} \\
Gemma-3-27B-it & 30.29 & \rqpair{\rqscore{\textbf{30.10}}{+9.0}}{\rqscore{27.62}{}} \\
\midrule
\textit{Advanced Models} &  & \rqtriple{{DeepSeek-V4.1-Flash}}{{DeepSeek-V4-Pro}}{{GPT-5.6-Sol}} \\
\cmidrule(lr){3-3}
DeepSeek-V4.1-Flash & 65.14 & \rqtriple{\rqscore{56.57}{}}{\rqscore{56.57}{0.0}}{\colorbox{black!6}{\rqscore{\textbf{72.95}}{+29.0}}} \\
DeepSeek-V4-Pro & 53.71 & \rqtriple{\rqscore{53.90}{+6.4}}{\rqscore{50.67}{}}{\rqscore{\textbf{72.19}}{+42.5}} \\
GPT-5.6-Sol & 65.71 & \rqtriple{\rqscore{64.95}{-2.8}}{\rqscore{60.00}{-10.3}}{\rqscore{\textbf{66.86}}{}} \\
\bottomrule
\end{tabular}
\end{table}

\begin{table}[!htbp]
\centering
\caption{Planner--coder pairing results on Code2Bench (Pass@1). Parentheses show the relative change from the same-model pair for each coder. Bold marks the best plan-assisted result for each coder; gray marks the highest-scoring pair in each model group.}
\label{tab:rq1-code2bench}
\footnotesize
\newcommand{\rqscore}[2]{\makebox[2.6em][r]{#1}\hspace{.25em}\makebox[4.1em][r]{\if\relax\detokenize{#2}\relax\else(#2\%)\fi}}
\setlength{\fboxsep}{1pt}
\renewcommand{\arraystretch}{1.15}
\newcommand{\rqpair}[2]{\makebox[.60\linewidth]{\makebox[.30\linewidth]{#1}\makebox[.30\linewidth]{#2}}}
\newcommand{\rqtriple}[3]{\makebox[.60\linewidth]{\makebox[.20\linewidth]{#1}\makebox[.20\linewidth]{#2}\makebox[.20\linewidth]{#3}}}
\begin{tabular}{@{}l@{\hspace{12pt}}r@{\hspace{12pt}}c@{}}
\toprule
\textbf{Coder} & \textbf{Direct} & \textbf{Planner} \\
\midrule
\textit{3B--4B} &  & \rqtriple{{Qwen3-4B}}{{Gemma-3-4B-it}}{{Ministral-3-3B-Instruct}} \\
\cmidrule(lr){3-3}
Qwen3-4B & 24.88 & \rqtriple{\rqscore{\textbf{22.58}}{}}{\rqscore{18.59}{-17.7}}{\rqscore{17.82}{-21.1}} \\
Gemma-3-4B-it & 21.20 & \rqtriple{\rqscore{\textbf{22.43}}{+23.7}}{\rqscore{18.13}{}}{\rqscore{16.59}{-8.5}} \\
Ministral-3-3B-Instruct & 24.88 & \rqtriple{\colorbox{black!6}{\rqscore{\textbf{23.96}}{+30.0}}}{\rqscore{19.97}{+8.3}}{\rqscore{18.43}{}} \\
\midrule
\textit{8B} &  & \rqpair{{Qwen3-8B}}{{Ministral-3-8B-Instruct}} \\
\cmidrule(lr){3-3}
Qwen3-8B & 24.88 & \rqpair{\rqscore{\textbf{27.04}}{}}{\rqscore{23.20}{-14.2}} \\
Ministral-3-8B-Instruct & 28.57 & \rqpair{\colorbox{black!6}{\rqscore{\textbf{27.96}}{+19.7}}}{\rqscore{23.35}{}} \\
\midrule
\textit{12B--14B} &  & \rqtriple{{Qwen3-14B}}{{Gemma-3-12B-it}}{{Ministral-3-14B-Instruct}} \\
\cmidrule(lr){3-3}
Qwen3-14B & 32.26 & \rqtriple{\colorbox{black!6}{\rqscore{\textbf{32.41}}{}}}{\rqscore{29.19}{-10.0}}{\rqscore{27.96}{-13.7}} \\
Gemma-3-12B-it & 28.57 & \rqtriple{\rqscore{\textbf{29.80}}{+18.3}}{\rqscore{25.19}{}}{\rqscore{26.11}{+3.7}} \\
Ministral-3-14B-Instruct & 30.41 & \rqtriple{\rqscore{\textbf{32.26}}{+14.1}}{\rqscore{29.65}{+4.9}}{\rqscore{28.26}{}} \\
\midrule
\textit{27B--32B} &  & \rqpair{{Qwen3-32B}}{{Gemma-3-27B-it}} \\
\cmidrule(lr){3-3}
Qwen3-32B & 29.49 & \rqpair{\rqscore{30.57}{}}{\colorbox{black!6}{\rqscore{\textbf{30.88}}{+1.0}}} \\
Gemma-3-27B-it & 30.88 & \rqpair{\rqscore{\textbf{29.95}}{+10.8}}{\rqscore{27.04}{}} \\
\midrule
\textit{Advanced Models} &  & \rqtriple{{DeepSeek-V4.1-Flash}}{{DeepSeek-V4-Pro}}{{GPT-5.6-Sol}} \\
\cmidrule(lr){3-3}
DeepSeek-V4.1-Flash & 36.41 & \rqtriple{\rqscore{35.48}{}}{\rqscore{36.56}{+3.0}}{\rqscore{\textbf{40.25}}{+13.4}} \\
DeepSeek-V4-Pro & 36.41 & \rqtriple{\rqscore{35.94}{-0.8}}{\rqscore{36.25}{}}{\rqscore{\textbf{40.25}}{+11.0}} \\
GPT-5.6-Sol & 36.87 & \rqtriple{\rqscore{36.10}{-12.3}}{\rqscore{36.56}{-11.2}}{\colorbox{black!6}{\rqscore{\textbf{41.17}}{}}} \\
\bottomrule
\end{tabular}
\end{table}

\textbf{Observation 1: Using the same LLM for planning and coding does not provide a systematic advantage.}
Across the two benchmarks, 27 of the 44 cross-model pairs (61.4\%) perform no worse than the corresponding same-model pair for the same coder, including 14 of 22 pairs on LiveCodeBench and 13 of 22 on Code2Bench. Moreover, the best-performing configuration is cross-model in seven of the ten model-group settings across the two benchmarks. These patterns are observed across different model scales. For example, on LiveCodeBench, using Qwen3-14B as the planner improves Gemma-3-12B-it and Ministral-3-14B-Instruct-2512 over their same-model results by 13.6\% and 9.0\%, respectively. Among the advanced models, using GPT-5.6-Sol as the planner improves DeepSeek-V4.1-Flash by 29.0\% and DeepSeek-V4-Pro by 42.5\%. Overall, the planning information seems to be a good intermediate representation to decouple planners and coders.

\textbf{Observation 2: The relative effectiveness of planners is largely consistent across different coders.}
In 9 of the 10 model-group settings across the two benchmarks, the same planner achieves the highest Pass@1 for every coder within the group. This pattern appears across different model scales and on both benchmarks. In the 3B--4B, 8B, and 12B--14B groups, Qwen3 consistently achieves the best planning results for all coders on both LiveCodeBench and Code2Bench. Among the advanced models, GPT-5.6-Sol similarly achieves the best results for all three coders on both benchmarks. The only exception occurs in the 27B--32B group on Code2Bench, where the preferred planner differs between the two coders. However, this difference is marginal for the Qwen3-32B coder, with Gemma-3-27B-it exceeding Qwen3-32B by only 0.31 Pass@1 (30.88 vs.\ 30.57), while Qwen3-32B retains a clear advantage for the Gemma-3-27B-it coder (29.95 vs.\ 27.04). Overall, the relative effectiveness of planners remains largely stable when paired with different coders.

\textbf{Observation 3: Strong planner-side and coder-side performance usually leads to the best configuration.}
At the model-group level, 7 of the 10 comparisons have a best-performing configuration that combines a planner consistently performing best across coders with a coder performing best or tied-best under most planners. The planner-side pattern appears in 9 of the 10 comparisons, as shown in Observation 2, while the corresponding coder-side pattern appears in 8 of the 10 comparisons. Their overlap in 7 comparisons indicates that the strongest configuration is usually explained by the effectiveness of its two components rather than by an isolated planner--coder match.

The 8B group provides a clear example on both benchmarks. Qwen3-8B achieves the best planning results for both coders, while Ministral-3-8B-Instruct-2512 achieves the best coding results under both planners. Their combination therefore produces the best 8B result on both LiveCodeBench (28.95) and Code2Bench (27.96). Note that this principle is irrespective of whether the same or different LLMs are used in the two stages. On LiveCodeBench, Qwen3-32B is strongest on both the planning and coding sides, and using Qwen3-32B for both stages gives the best 27B--32B result at 31.62. Among the seven comparisons following this pattern, five are from a cross-model configuration and two are from a same-model configuration.

Direct performance provides an additional reference for coding effectiveness. In 6 of the 10 model-group comparisons, the coder in the best configuration also has the highest or tied-highest direct Pass@1. For example, Ministral-3-8B-Instruct-2512 has the highest Direct performance in the 8B group and also remains the stronger coder under planning on both benchmarks. However, this correspondence is not universal, as shown by the 27B--32B group on LiveCodeBench, where Gemma-3-27B-it has higher Direct performance but Qwen3-32B performs better as the coder under either planner. 

Overall, our results suggest that strong configurations can be obtained by considering planner-side and coder-side effectiveness separately, rather than relying on a specific relationship between model identities.

\begin{findingbox}
\textit{Finding 1.}
Cross-model planner--coder configurations are generally competitive with same-model configurations, and pairing stronger planners with stronger coders tends to yield better combined performance. This finding supports combining an effective planner with a strong coder across model boundaries, rather than restricting both stages to the same LLM.
\end{findingbox}

\subsection{RQ2: Scale Requirements of Planning and Coding}

\paragraph{Design.}
We examine model scaling from both the planner and coder sides. For planner scaling, we vary model sizes within the Qwen3, Gemma-3, and Ministral-3 families with fixed small and larger coders, and further include Qwen3.5-27B and three advanced models to broaden the planning settings. For coder scaling, Qwen3.5-27B and the three advanced models  are paired with different coder sizes from the same three families, together with Qwen3.5-27B and Qwen3.8-27B. By varying one side while fixing the other, we examine the scaling behavior of each stage and whether planning and coding require comparable model scale.

\begin{figure}[!htbp]
\centering
\includegraphics[width=\linewidth]{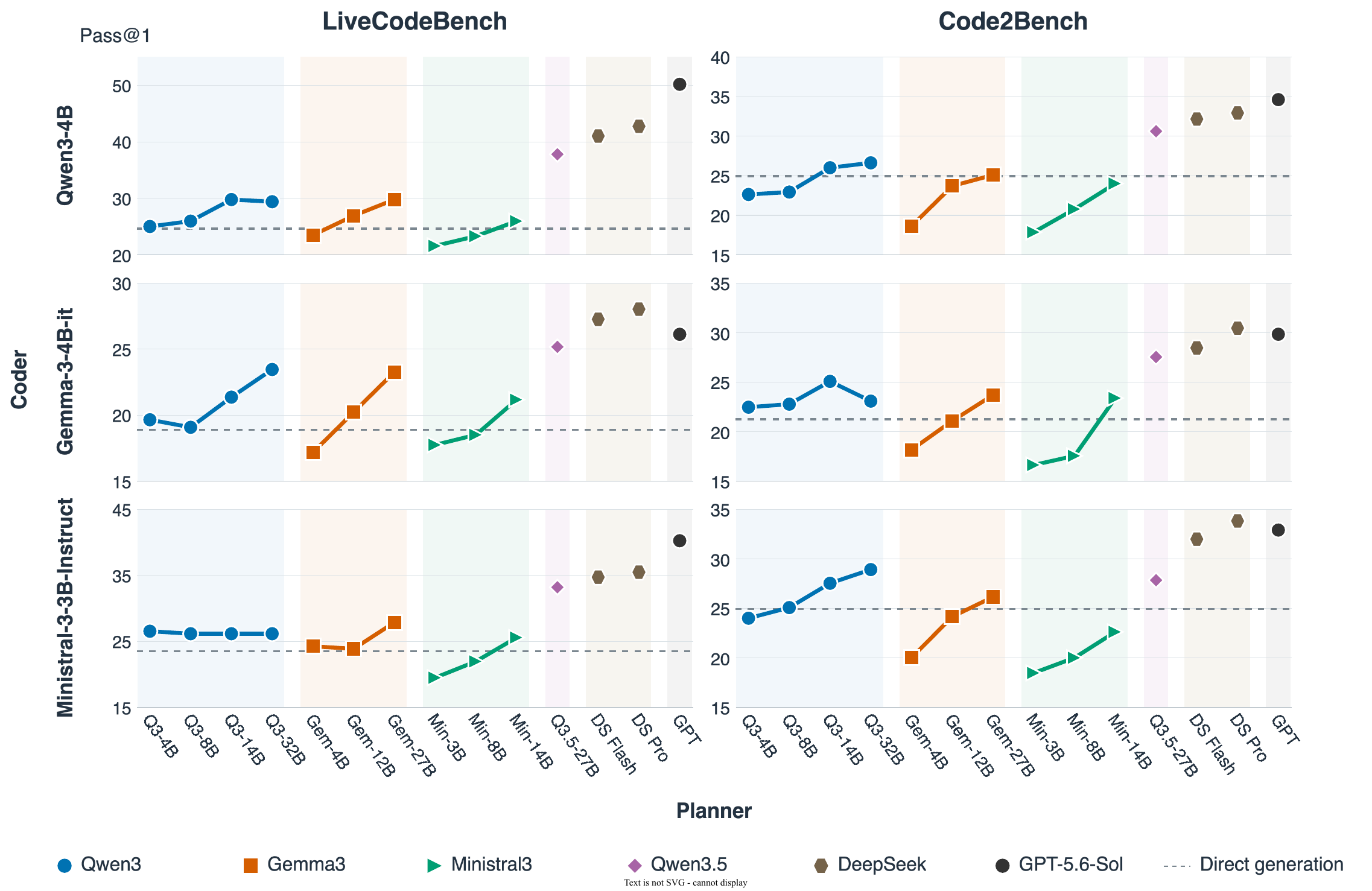}
\caption{Planner performance across model scales with three small coders. Q3, Gem, and Min abbreviate Qwen3, Gemma-3-it, and Ministral-3-Instruct-2512; DS Flash/Pro abbreviate DeepSeek-V4.1-Flash/V4-Pro, and GPT abbreviates GPT-5.6-Sol.}
\label{fig:rq2-planner-small-coders}
\Description{Six panels compare planner models with fixed Qwen3-4B, Gemma-3-4B-it, and Ministral-3-3B-Instruct-2512 coders, arranged in three rows. The left column shows LiveCodeBench and the right column shows Code2Bench. The horizontal axis lists planners and the vertical axis shows Pass@1 percentages. Lines connect Qwen3, Gemma-3, and Ministral-3 planners of different sizes. Separate points show Qwen3.5-27B, DeepSeek-V4.1-Flash, DeepSeek-V4-Pro, and GPT-5.6-Sol. Dashed horizontal lines mark direct generation by each coder. Increasing planner size generally improves small-coder performance, with exceptions among Qwen3 pairings.}
\end{figure}

\begin{figure}[!htbp]
\centering
\includegraphics[width=\linewidth]{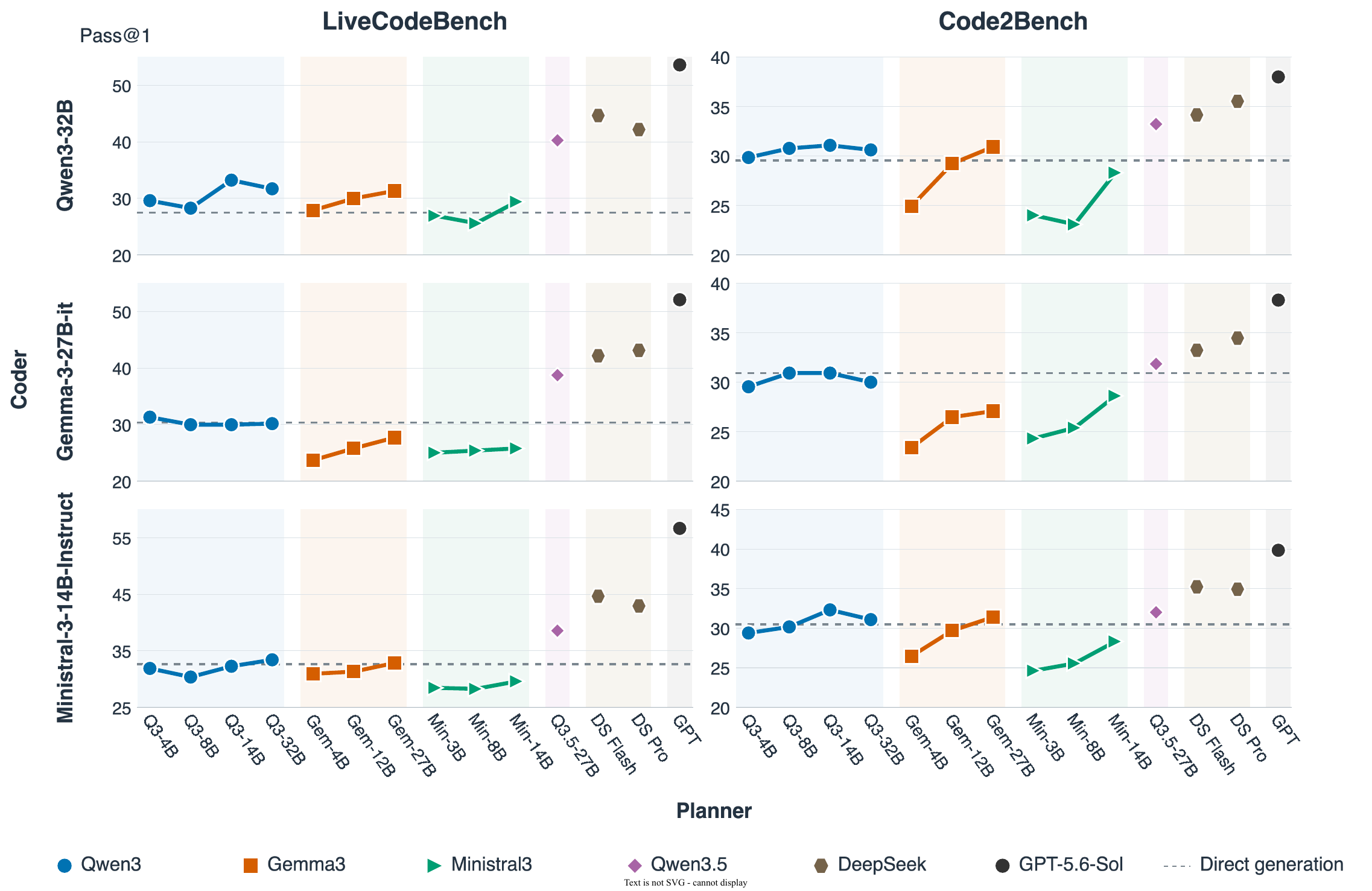}
\caption{Planner performance across model scales with three larger coders. Q3, Gem, and Min abbreviate Qwen3, Gemma-3-it, and Ministral-3-Instruct-2512; DS Flash/Pro abbreviate DeepSeek-V4.1-Flash/V4-Pro, and GPT abbreviates GPT-5.6-Sol.}
\label{fig:rq2-planner-large-coders}
\Description{Six panels compare planner models with fixed Qwen3-32B, Gemma-3-27B-it, and Ministral-3-14B-Instruct-2512 coders, arranged in three rows. The left column shows LiveCodeBench and the right column shows Code2Bench. The horizontal axis lists planners and the vertical axis shows Pass@1 percentages. Lines connect Qwen3, Gemma-3, and Ministral-3 planners of different sizes. Separate points show Qwen3.5-27B, DeepSeek-V4.1-Flash, DeepSeek-V4-Pro, and GPT-5.6-Sol. Dashed horizontal lines mark direct generation by each coder. Within-family scaling trends vary by coder; Qwen3.5 and the advanced planners exceed the direct baselines in all six panels.}
\end{figure}

\paragraph{Results.}
Figures~\ref{fig:rq2-planner-small-coders} and~\ref{fig:rq2-planner-large-coders} show how performance changes with planner scale for small and larger coders, while Tables~\ref{tab:rq2-lcb} and~\ref{tab:rq2-code2bench} show how performance changes with coder scale under different planners. These results lead to the following observations.

\begin{table}[!htbp]
\centering
\caption{Coder performance across model scales with four planners on LiveCodeBench (Pass@1). Parentheses show the relative change from Direct for each coder. Bold marks the best planner for each coder; gray marks the best coder for each planner within each group.}
\label{tab:rq2-lcb}
\footnotesize
\newcommand{\rqscore}[2]{\makebox[2.6em][r]{#1}\hspace{.25em}{\makebox[4.1em][r]{(#2\%)}}}
\setlength{\fboxsep}{1pt}
\renewcommand{\arraystretch}{1.15}
\setlength{\tabcolsep}{4pt}
\begin{tabular}{@{}lrcccc@{}}
\toprule
\multirow{2}{*}{\textbf{Coder}} & \multirow{2}{*}{\textbf{Direct}} & \multicolumn{4}{c}{\textbf{Planner}} \\
\cmidrule(lr){3-6}
 & & Qwen3.5-27B & DeepSeek-V4.1-Flash & DeepSeek-V4-Pro & GPT-5.6-Sol \\
\midrule
Qwen3-4B & 24.57 & \rqscore{37.71}{+53.5} & \rqscore{40.95}{+66.7} & \rqscore{42.67}{+73.6} & \rqscore{\textbf{50.10}}{+103.9} \\
Qwen3-8B & 24.57 & \rqscore{39.05}{+58.9} & \rqscore{42.29}{+72.1} & \rqscore{43.24}{+76.0} & \rqscore{\textbf{51.24}}{+108.5} \\
Qwen3-14B & 28.00 & \colorbox{black!6}{\rqscore{40.57}{+44.9}} & \rqscore{44.38}{+58.5} & \colorbox{black!6}{\rqscore{46.86}{+67.3}} & \colorbox{black!6}{\rqscore{\textbf{55.24}}{+97.3}} \\
Qwen3-32B & 27.43 & \rqscore{40.19}{+46.5} & \colorbox{black!6}{\rqscore{44.57}{+62.5}} & \rqscore{42.10}{+53.5} & \rqscore{\textbf{53.52}}{+95.1} \\
\midrule
Gemma-3-4B-it & 18.86 & \rqscore{25.14}{+33.3} & \rqscore{27.24}{+44.4} & \rqscore{\textbf{28.00}}{+48.5} & \rqscore{26.10}{+38.4} \\
Gemma-3-12B-it & 25.71 & \rqscore{36.38}{+41.5} & \rqscore{38.48}{+49.6} & \rqscore{38.10}{+48.1} & \rqscore{\textbf{44.76}}{+74.1} \\
Gemma-3-27B-it & 30.29 & \colorbox{black!6}{\rqscore{38.67}{+27.7}} & \colorbox{black!6}{\rqscore{42.10}{+39.0}} & \colorbox{black!6}{\rqscore{43.05}{+42.1}} & \colorbox{black!6}{\rqscore{\textbf{52.00}}{+71.7}} \\
\midrule
Ministral-3-3B-Instruct & 23.43 & \rqscore{33.14}{+41.5} & \rqscore{34.67}{+48.0} & \rqscore{35.43}{+51.2} & \rqscore{\textbf{40.19}}{+71.5} \\
Ministral-3-8B-Instruct & 29.14 & \rqscore{38.29}{+31.4} & \rqscore{40.76}{+39.9} & \rqscore{42.67}{+46.4} & \rqscore{\textbf{52.38}}{+79.7} \\
Ministral-3-14B-Instruct & 32.57 & \colorbox{black!6}{\rqscore{38.48}{+18.1}} & \colorbox{black!6}{\rqscore{44.57}{+36.8}} & \colorbox{black!6}{\rqscore{42.86}{+31.6}} & \colorbox{black!6}{\rqscore{\textbf{56.57}}{+73.7}} \\
\midrule
Qwen3.5-27B & 46.86 & \rqscore{48.19}{+2.8} & \rqscore{55.81}{+19.1} & \rqscore{52.00}{+11.0} & \rqscore{\textbf{68.00}}{+45.1} \\
Qwen3.8-27B & 58.29 & \colorbox{black!6}{\rqscore{50.86}{-12.7}} & \colorbox{black!6}{\rqscore{59.43}{+2.0}} & \colorbox{black!6}{\rqscore{55.24}{-5.2}} & \colorbox{black!6}{\rqscore{\textbf{71.05}}{+21.9}} \\
\bottomrule
\end{tabular}
\end{table}

\begin{table}[!htbp]
\centering
\caption{Coder performance across model scales with four planners on Code2Bench (Pass@1). Parentheses show the relative change from Direct for each coder. Bold marks the best planner for each coder; gray marks the best coder for each planner within each group.}
\label{tab:rq2-code2bench}
\footnotesize
\newcommand{\rqscore}[2]{\makebox[2.6em][r]{#1}\hspace{.25em}{\makebox[4.1em][r]{(#2\%)}}}
\setlength{\fboxsep}{1pt}
\renewcommand{\arraystretch}{1.15}
\setlength{\tabcolsep}{4pt}
\begin{tabular}{@{}lrcccc@{}}
\toprule
\multirow{2}{*}{\textbf{Coder}} & \multirow{2}{*}{\textbf{Direct}} & \multicolumn{4}{c}{\textbf{Planner}} \\
\cmidrule(lr){3-6}
 & & Qwen3.5-27B & DeepSeek-V4.1-Flash & DeepSeek-V4-Pro & GPT-5.6-Sol \\
\midrule
Qwen3-4B & 24.88 & \rqscore{30.57}{+22.8} & \rqscore{32.10}{+29.0} & \rqscore{32.87}{+32.1} & \rqscore{\textbf{34.56}}{+38.9} \\
Qwen3-8B & 24.88 & \rqscore{31.80}{+27.8} & \rqscore{32.41}{+30.2} & \rqscore{34.25}{+37.7} & \rqscore{\textbf{34.87}}{+40.1} \\
Qwen3-14B & 32.26 & \colorbox{black!6}{\rqscore{33.64}{+4.3}} & \rqscore{33.95}{+5.2} & \rqscore{35.02}{+8.6} & \colorbox{black!6}{\rqscore{\textbf{37.94}}{+17.6}} \\
Qwen3-32B & 29.49 & \rqscore{33.18}{+12.5} & \colorbox{black!6}{\rqscore{34.10}{+15.6}} & \colorbox{black!6}{\rqscore{35.48}{+20.3}} & \colorbox{black!6}{\rqscore{\textbf{37.94}}{+28.6}} \\
\midrule
Gemma-3-4B-it & 21.20 & \rqscore{27.50}{+29.7} & \rqscore{28.42}{+34.1} & \rqscore{\textbf{30.41}}{+43.5} & \rqscore{29.80}{+40.6} \\
Gemma-3-12B-it & 28.57 & \rqscore{31.34}{+9.7} & \colorbox{black!6}{\rqscore{33.79}{+18.3}} & \colorbox{black!6}{\rqscore{34.72}{+21.5}} & \rqscore{\textbf{36.41}}{+27.4} \\
Gemma-3-27B-it & 30.88 & \colorbox{black!6}{\rqscore{31.80}{+3.0}} & \rqscore{33.18}{+7.5} & \rqscore{34.41}{+11.4} & \colorbox{black!6}{\rqscore{\textbf{38.25}}{+23.9}} \\
\midrule
Ministral-3-3B-Instruct & 24.88 & \rqscore{27.80}{+11.7} & \rqscore{31.95}{+28.4} & \rqscore{\textbf{33.79}}{+35.8} & \rqscore{32.87}{+32.1} \\
Ministral-3-8B-Instruct & 28.57 & \colorbox{black!6}{\rqscore{32.57}{+14.0}} & \rqscore{33.95}{+18.8} & \colorbox{black!6}{\rqscore{35.18}{+23.1}} & \rqscore{\textbf{35.79}}{+25.3} \\
Ministral-3-14B-Instruct & 30.41 & \rqscore{31.95}{+5.1} & \colorbox{black!6}{\rqscore{35.18}{+15.7}} & \rqscore{34.87}{+14.6} & \colorbox{black!6}{\rqscore{\textbf{39.78}}{+30.8}} \\
\midrule
Qwen3.5-27B & 34.10 & \rqscore{34.56}{+1.4} & \colorbox{black!6}{\rqscore{34.56}{+1.4}} & \rqscore{35.94}{+5.4} & \rqscore{\textbf{38.10}}{+11.7} \\
Qwen3.8-27B & 34.56 & \colorbox{black!6}{\rqscore{35.02}{+1.3}} & \rqscore{34.10}{-1.3} & \colorbox{black!6}{\rqscore{36.25}{+4.9}} & \colorbox{black!6}{\rqscore{\textbf{38.40}}{+11.1}} \\
\bottomrule
\end{tabular}
\end{table}

\textbf{Observation 4: Increasing planner scale generally improves plan-assisted performance.}
With small coders fixed, plan-assisted performance generally increases with planner scale within the same model family, particularly for Gemma-3 and Ministral-3. For example, with Qwen3-4B as the coder, scaling the Gemma-3 planner from 4B to 27B improves Pass@1 from 23.43 to 29.71 on LiveCodeBench and from 18.59 to 25.04 on Code2Bench. Qwen3 shows a less pronounced scaling trend, as its smaller variants already provide relatively effective plans, consistent with the observation in RQ1. Qwen3.5-27B and the advanced models achieve further improvements, showing that better planner-side performance can provide additional gains.

\textbf{Observation 5: Strong planners consistently benefit larger coders.}
When coder scale is relatively large, plans from smaller models provide limited and less consistent gains over direct generation. This may be because larger coders can already obtain much of the required solution information directly from the task, reducing the additional value of plans from smaller models. In contrast, strong planners such as Qwen3.5-27B and the advanced models consistently improve all three larger coders on both benchmarks. For example, with Qwen3-32B as the coder, several smaller planners remain close to its direct performance of 27.43 and 29.49 on LiveCodeBench and Code2Bench, whereas GPT-5.6-Sol increases Pass@1 to 53.52 and 37.94. This contrast suggests that scaling up the coder does not eliminate the advantage of retaining a large model for planning.

\textbf{Observation 6: Increasing coder scale generally improves plan-assisted performance.}
When the planner is fixed, plan-assisted performance generally increases with the coder scale, particularly for the Gemma-3 and Ministral-3 families. For example, with GPT-5.6-Sol as the planner on LiveCodeBench, scaling Gemma-3 from 4B to 27B improves Pass@1 by 25.90 points, while scaling Ministral-3 from 3B to 14B improves it by 16.38 points. Qwen3 follows the same trend up to 14B, but its 32B variant performs slightly below 14B (53.52 vs.\ 55.24). This reversal is also observed in direct generation on both benchmarks, suggesting that it may be due to specific characteristics of the two models. One possible explanation is the difference in post-training: Qwen3 models up to 14B use Strong-to-Weak Distillation, whereas Qwen3-32B follows a separate four-stage post-training process~\cite{yang2025qwen3}. Overall, the results show a clear general benefit from increasing coder scale, with Qwen3-32B as a model-specific exception.

\textbf{Observation 7: Planning and coding do not require comparable model scale.}
A strong planner can retain nearly the same effectiveness when paired with a substantially smaller coder. With GPT-5.6-Sol as the planner, the two 27B coders already achieve performance comparable to the advanced coders on both benchmarks. On LiveCodeBench, Qwen3.5-27B and Qwen3.8-27B achieve 68.00 and 71.05, respectively, compared with 66.86--72.95 for the three advanced coders in Table~\ref{tab:rq1-lcb}. On Code2Bench, they achieve 38.10 and 38.40, compared with 40.25--41.17 for the advanced coders in Table~\ref{tab:rq1-code2bench}. Therefore, using an advanced model for coding provides only limited additional benefit over a 27B coder when the same strong planner is used.

In some settings, the coding model can be reduced further while remaining competitive. On Code2Bench, GPT-5.6-Sol paired with Ministral-3-14B-Instruct achieves 39.78, only 0.47 percentage points of Pass@1 below either DeepSeek advanced coder and 1.39 percentage points below GPT-5.6-Sol itself as the coder. More aggressive scale reduction can still preserve substantial gains from planning: Qwen3.5-27B paired with Qwen3-4B achieves 37.71 on LiveCodeBench and 30.57 on Code2Bench, outperforming the direct performance of the eight-times-larger Qwen3-32B on both benchmarks (27.43 and 29.49). These results demonstrate that effective planning can substantially reduce the model capacity required on the coder side.

These results reveal asymmetric scaling behaviors between planning and coding. Increasing planner scale generally continues to improve performance across coder scales, whereas the gains from increasing coder scale become more limited under a strong planner. As a result, the coder does not need to scale to the same extent as the planner. Under a strong planner, substantially smaller coders can approach the performance of much stronger coders and, in some settings, remain competitive with using a large model for both planning and coding. This asymmetry motivates Large-Planner-Small-Coder (LPSC), in which a large model is retained for planning while the coding stage uses a substantially smaller model.

\begin{findingbox}
\textit{Finding 2.}
Planning and coding do not require comparable model scales. A large planner paired with a substantially smaller coder can sometimes remain competitive with using the large model for both stages, supporting LPSC as a practical configuration that retains a large model for planning while substantially reducing the model scale used for coding.
\end{findingbox}

\subsection{RQ3: Transferability of Coder Improvement}

\paragraph{Design.}
We further train three coders: Qwen2.5-Coder-1.5B-Instruct, Qwen2.5-Coder-3B-Instruct, and Gemma-3-4B-it. For each coder, we compare its original and fine-tuned versions under both direct and plan-assisted generation. For plan-assisted generation, we reuse the same plans from Qwen3.5-27B and advanced models as in RQ2. For each planner, the original and fine-tuned coders are evaluated using exactly the same plans.

\paragraph{Results.}
Tables~\ref{tab:rq3-lcb} and~\ref{tab:rq3-code2bench} present comparison results of coder performance before and after SFT across different planners. The results reveal the following observations.

\begin{table}[!htbp]
\centering
\caption{Base and SFT coder performance on LiveCodeBench (Pass@1). Relative Change is computed from Base to SFT. Bold marks the higher Base/SFT score for each condition.}
\label{tab:rq3-lcb}
\footnotesize
\setlength{\tabcolsep}{3pt}
\renewcommand{\arraystretch}{1.15}
\begin{tabular}{@{}llrcccc@{}}
\toprule
\multirow{2}{*}{\textbf{Coder}} & \multirow{2}{*}{\textbf{Setting}} & \multirow{2}{*}{\textbf{Direct}} & \multicolumn{4}{c}{\textbf{Planner}} \\
\cmidrule(lr){4-7}
 & & & Qwen3.5-27B & DeepSeek-V4.1-Flash & DeepSeek-V4-Pro & GPT-5.6-Sol \\
\midrule
\multirow{2}{*}{Qwen2.5-Coder-1.5B} & Base & 10.86 & 23.05 & \textbf{24.95} & 25.90 & 28.00 \\
 & SFT & \textbf{13.71} & \textbf{24.19} & \textbf{24.95} & \textbf{27.43} & \textbf{28.95} \\
\cmidrule(lr){1-7}
\multicolumn{2}{c}{Relative Change (\%)} & +26.3 & +5.0 & +0.0 & +5.9 & +3.4 \\
\midrule
\multirow{2}{*}{Qwen2.5-Coder-3B} & Base & 14.29 & 26.67 & 29.14 & 32.95 & 34.48 \\
 & SFT & \textbf{20.00} & \textbf{29.14} & \textbf{32.19} & \textbf{34.48} & \textbf{37.33} \\
\cmidrule(lr){1-7}
\multicolumn{2}{c}{Relative Change (\%)} & +40.0 & +9.3 & +10.5 & +4.6 & +8.3 \\
\midrule
\multirow{2}{*}{Gemma-3-4B-it} & Base & \textbf{18.86} & 25.14 & \textbf{27.24} & \textbf{28.00} & \textbf{26.10} \\
 & SFT & 13.71 & \textbf{25.52} & 24.19 & 26.29 & 25.71 \\
\cmidrule(lr){1-7}
\multicolumn{2}{c}{Relative Change (\%)} & -27.3 & +1.5 & -11.2 & -6.1 & -1.5 \\
\bottomrule
\end{tabular}
\end{table}

\begin{table}[!htbp]
\centering
\caption{Base and SFT coder performance on Code2Bench (Pass@1). Relative Change (\%) is computed from Base to SFT. Bold marks the higher Base/SFT score for each condition.}
\label{tab:rq3-code2bench}
\footnotesize
\setlength{\tabcolsep}{3pt}
\renewcommand{\arraystretch}{1.15}
\begin{tabular}{@{}llrcccc@{}}
\toprule
\multirow{2}{*}{\textbf{Coder}} & \multirow{2}{*}{\textbf{Setting}} & \multirow{2}{*}{\textbf{Direct}} & \multicolumn{4}{c}{\textbf{Planner}} \\
\cmidrule(lr){4-7}
 & & & Qwen3.5-27B & DeepSeek-V4.1-Flash & DeepSeek-V4-Pro & GPT-5.6-Sol \\
\midrule
\multirow{2}{*}{Qwen2.5-Coder-1.5B} & Base & 15.21 & 25.19 & 25.65 & 25.96 & 23.35 \\
 & SFT & \textbf{21.20} & \textbf{27.65} & \textbf{29.19} & \textbf{31.18} & \textbf{30.26} \\
\cmidrule(lr){1-7}
\multicolumn{2}{c}{Relative Change (\%)} & +39.4 & +9.8 & +13.8 & +20.1 & +29.6 \\
\midrule
\multirow{2}{*}{Qwen2.5-Coder-3B} & Base & 20.28 & 28.11 & 27.96 & 29.95 & 30.57 \\
 & SFT & \textbf{25.81} & \textbf{29.49} & \textbf{31.49} & \textbf{31.64} & \textbf{34.10} \\
\cmidrule(lr){1-7}
\multicolumn{2}{c}{Relative Change (\%)} & +27.3 & +4.9 & +12.6 & +5.6 & +11.6 \\
\midrule
\multirow{2}{*}{Gemma-3-4B-it} & Base & 21.20 & 27.50 & 28.42 & 30.41 & 29.80 \\
 & SFT & \textbf{25.81} & \textbf{28.88} & \textbf{29.65} & \textbf{32.10} & \textbf{31.80} \\
\cmidrule(lr){1-7}
\multicolumn{2}{c}{Relative Change (\%)} & +21.7 & +5.0 & +4.3 & +5.6 & +6.7 \\
\bottomrule
\end{tabular}
\end{table}

\textbf{Observation 8: Coder-side improvements consistently transfer across different planners.}
SFT improves direct generation in five of the six coder--benchmark settings. For these five settings, the same positive effect appears in 19 of the 20 corresponding plan-assisted settings, with the remaining one unchanged. For example, on LiveCodeBench, Qwen2.5-Coder-3B improves from 14.29 to 20.00 under direct generation, while its performance also increases under all four planners, with relative gains ranging from 4.6\% to 10.5\%. The pattern is even more consistent on Code2Bench, where all 12 plan-assisted settings benefit from SFT. Importantly, SFT uses only standard code instruction data, and the planning inputs are held fixed before and after training. The results therefore show that coder-side improvements learned independently of planning can be retained across plans generated by different LLMs without planner-specific adaptation.

\textbf{Observation 9: The effect of coder training can vary across benchmarks, but is largely preserved across planners.}
Gemma-3-4B-it provides a clear contrast. On LiveCodeBench, SFT reduces its direct performance from 18.86 to 13.71. Its performance also decreases under three of the four planners, by 11.2\%, 6.1\%, and 1.5\%, while the remaining planner yields only a marginal 1.5\% increase. In contrast, on Code2Bench, the same SFT improves direct performance from 21.20 to 25.81, and performance improves under all four planners, by 4.3\%--6.7\%. More generally, among all 24 plan-assisted Base--SFT comparisons, 22 change in the same direction as the corresponding direct result, one remains unchanged, and only one changes in the opposite direction. The direction of the training effect is therefore more closely associated with the coder--benchmark setting than with a particular planner source, indicating that changing the planner rarely reverses the effect of coder-side training.

The consistency of this transfer across the evaluated coder scales and model families points to a broader opportunity for planner-agnostic coder optimization. Although our training experiments are limited to relatively small coders due to resource constraints, this result is particularly relevant to LPSC: training effort can be concentrated on improving the small coder while retaining an already capable large planner. More effective coder-side training may therefore provide a direct path toward further strengthening LPSC and narrowing the gap to configurations that use large models for both planning and coding.

\begin{findingbox}
\textit{Finding 3.}
Further training of the coder generally improves its performance across different planners, with the resulting gains largely preserved across plan sources. This supports planner-agnostic coder-side optimization, allowing training to focus on the coder itself without being tailored to a particular planner and providing a direct path for further strengthening LPSC through targeted improvement of the small coder.
\end{findingbox}

\section{Case Study}

\begin{figure*}[t]
    \centering
    \includegraphics[width=\textwidth]{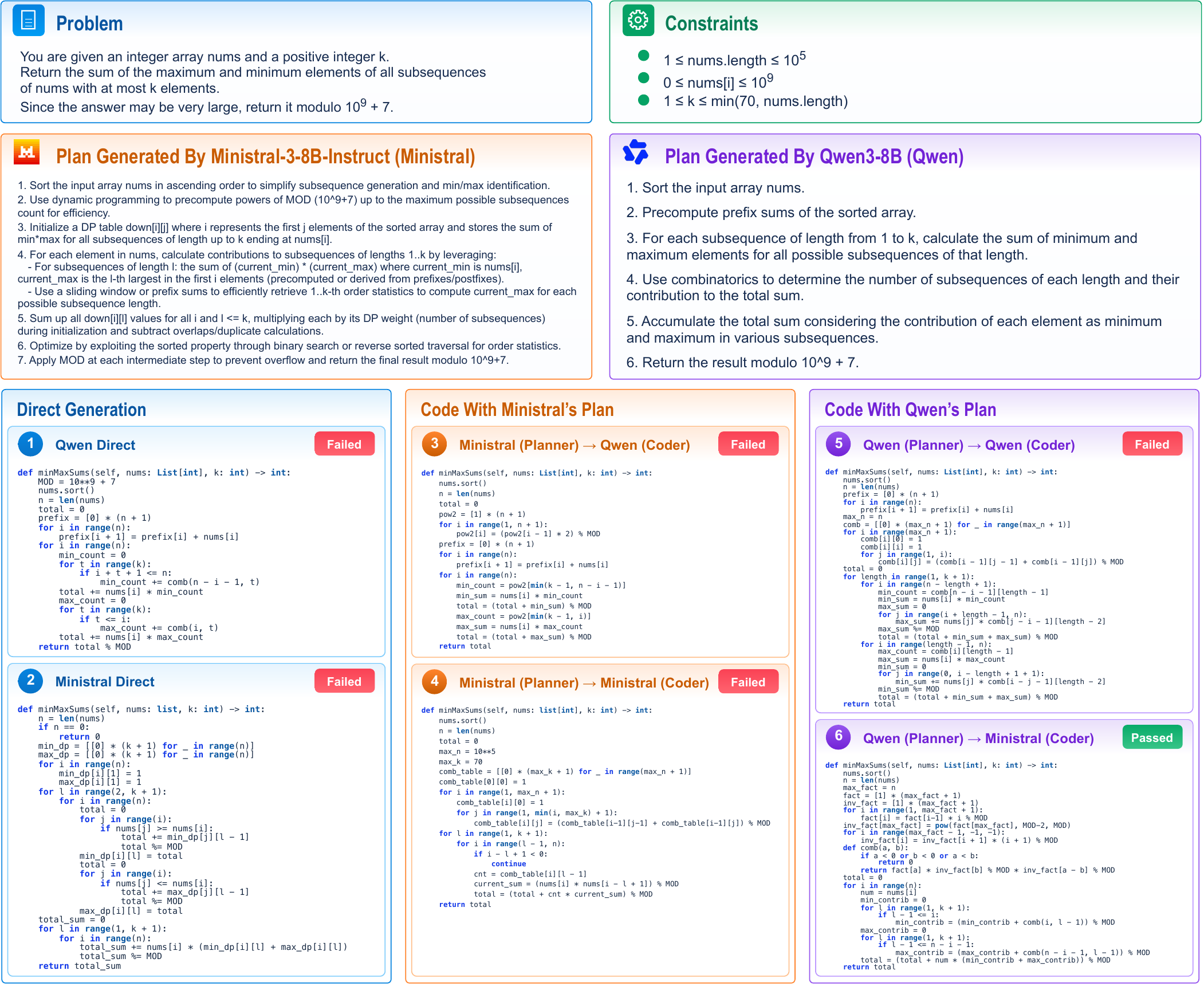}
    \caption{Plans and generated code from Qwen3-8B and Ministral-3-8B-Instruct for LiveCodeBench problem 3714, comparing direct generation with four planner-coder pairings.}
    \label{fig:case-study}
    \Description{The problem, input constraints, two generated plans, and six code-generation outcomes: both direct generations fail, and only Qwen3-8B planning paired with Ministral-3-8B-Instruct coding passes.}
\end{figure*}

To further understand the behaviors observed in our evaluation, we analyze a representative cross-model case involving Qwen3-8B and Ministral-3-8B-Instruct. As shown in Figure~\ref{fig:case-study}, both models fail when directly generating code for this problem. Furthermore, only the configuration using Qwen3-8B as the planner and Ministral-3-8B-Instruct as the coder produces a correct solution; all other planner-coder pairs fail to generate a correct solution.

\textbf{A model with failed direct generation can still produce effective solution plans.}
Although Qwen3-8B fails to generate correct code directly, the solution plan produced by Qwen3-8B provides the key strategy used by the successful planner-coder pair. The plan proposes sorting the array and using combinatorial counting to compute how each element contributes when it serves as the minimum or maximum of subsequences with at most \(k\) elements. This strategy avoids enumerating all possible subsequences and provides useful guidance for implementation. This observation suggests that direct code generation performance does not fully reflect a model's ability to provide useful solution plans.

\textbf{The generated plan alone is insufficient for successful code generation.}
The different outcomes of Qwen3-8B and Ministral-3-8B-Instruct as coders further explain the success of the cross-model pair. When Qwen3-8B consumes its own plan, the generated code remains incorrect. In contrast, Ministral-3-8B-Instruct turns the same plan into a correct implementation by precomputing factorials and modular inverses, calculating the contribution counts for each valid subsequence length, and accumulating the weighted contributions modulo \(10^9+7\). This comparison indicates that identifying a solution strategy and realizing it in executable code may not necessarily be achieved by the same model.

\textbf{Direct code generation performance does not fully characterize planner behavior.}
Ministral-3-8B-Instruct demonstrates stronger direct code generation performance than Qwen3-8B in our evaluation. However, assigning it as the planner does not produce the same outcome in this case. Its generated solution plan follows a dynamic programming-based approach with additional state definitions and optimization procedures, but this approach does not produce a correct solution. In contrast, the plan produced by Qwen3-8B is successfully used to generate a correct solution. These results suggest that direct code generation performance alone does not fully characterize a model's behavior when assigned as a planner.

This case illustrates why different planner-coder combinations can exhibit different behaviors. A model that fails to generate the final code may still provide useful solution plans, while another model may be better at realizing those plans.
\section{Threats to Validity}

\subsection{Construct Validity}
A potential threat is whether benchmark-based evaluation provides a reliable measurement of planner-coder interaction. Our evaluation uses functional correctness as the primary criterion, which is determined by the testing mechanism of each benchmark. LiveCodeBench evaluates generated programs against predefined test cases, where finite test coverage may miss some valid input scenarios. Code2Bench adopts property-based testing to generate diverse inputs and evaluates generated functions on the resulting test suites. However, the test generation process may introduce randomness in our evaluation setup, potentially causing the generated test inputs and correctness judgments to vary across runs.

To ensure a consistent evaluation, all planner-coder configurations are evaluated on the same benchmark tasks following the original evaluation pipelines of each benchmark, including their test data, generators, and runners without modification. In addition, evaluating on two complementary benchmarks reduces the dependence on any single testing methodology. Nevertheless, finite test coverage and potential randomness in automated testing remain inherent limitations of benchmark-based code generation evaluation.

\subsection{Internal Validity}

Generation variability may affect the measured performance of a planner-coder pair. This variability can arise in both stages. Plan generation uses sampling, so the same planner can produce different plans for the same problem, which may lead to different coding outcomes. For code generation, we use greedy decoding to reduce randomness, but complete determinism is still difficult to guarantee. For API-based models, some serving details are outside our control, while local inference with vLLM may also show small run-to-run differences.

We reduce the influence of these factors through a consistent evaluation protocol. For each planner and problem, we generate three independent plans, retain all of them, and average the corresponding coding results rather than selecting a favorable plan. The same plans are reused when comparing different coders, keeping the planning input fixed across coder comparisons. We also use greedy decoding throughout code generation to reduce additional variation.

\subsection{External Validity}

Our evaluation is based on two Python code-generation benchmarks and models from several families and scales. The observed patterns may not directly extend to other programming languages or other software-engineering tasks, such as repository-level development, debugging, or program repair. We also study a specific form of planner-coder interaction, where a natural-language solution plan is generated once and then passed to the coder. Workflows with different plan representations, execution feedback, plan revision, or repeated interaction may lead to different results.

Our coder-side training experiments cover three relatively small coders using standard code instruction tuning. The observed transfer of coder improvements across planners is therefore established within this training setting, and may differ for larger coders or substantially different training approaches.
\section{Related Work}

\subsection{Planning as Reasoning}

Planning has been explored as an explicit reasoning stage in code generation, where models first formulate intermediate solution information and then use it to guide code generation.  Inspired by Chain-of-Thought~(CoT) reasoning, Self-Planning~\cite{jiang2024selfplanning} introduces explicit solution steps before implementation, while other studies explore different forms of planning information, including structured solution representations~\cite{li2025structured}, hierarchical decomposition~\cite{zelikman2023parsel} and modular problem decomposition~\cite{pan2026mot}.

Beyond generating a single plan, later work also explores multiple candidate solutions to improve planning. PlanSearch~\cite{wang2025planning} searches over diverse natural-language plans, while other approaches explore alternative solution paths~\cite{gao2026seer}, identify consensus across candidate plans~\cite{wen2026consensus}, or organize candidate solutions through tree-structured exploration~\cite{li2025codetree,cui2026dualphase}. Empirical studies~\cite{zhang2026aretheygood,jin2027understanding} further examine the effectiveness of intermediate solution information, showing that its quality does not always align with the correctness of generated code and that different forms of intermediate information can lead to different generation outcomes.

This line of work mainly investigates how to generate effective plans and how planning information affects code generation. Our work shifts the focus from the quality of individual plans to the capabilities of models that generate and utilize these plans.

\subsection{Planning as Interface}

Plans can also serve as explicit interfaces between solution formulation and subsequent stages of code generation. Multi-agent approaches such as MapCoder~\cite{islam-etal-2024-mapcoder} separate planning from implementation and pass the generated plan to a coding stage. Other systems organize planning, coding, execution, and verification into different roles or components~\cite{lin2025soen,dong2024selfcollaboration,xu-etal-2026-mavencoder}.

Later work extends this basic plan-then-code process with iterative interaction. Some approaches verify or revise plans before implementation~\cite{lei-etal-2025-planning}, while others use simulation~\cite{islam-etal-2025-codesim} or iterative planning-coding collaboration~\cite{doan-etal-2026-collabcoder} to refine the solution. Other systems adapt the use of planning according to generation outcomes ~\cite{zhu2026adacoder,lee2026solidcoder}.

Recent studies further broaden how such interfaces are used. A study~\cite{hadhoud-etal-2026-idea} uses natural-language editorials to separate problem solving from implementation in code generation. Other work explores cross-model collaboration for practical objectives: COPE~\cite{lee2026efficient} uses plans to coordinate small and large LLMs for efficient inference, while PaT~\cite{yoon-etal-2026-pat} introduces planning after an unsuccessful initial generation to reduce unnecessary planning cost.

These studies demonstrate the usefulness of plans as interfaces for organizing code generation workflows. However, such interfaces are typically embedded in specific systems or designed for particular objectives. Our work also treats plans as interfaces. However, our work does not use specific plans for a fixed workflow, but studies how various plans can be used across various planner-coder configurations.

\subsection{Planning in Model Training}

Planning-related information has also been incorporated into the training of code generation models. Inspired by reasoning distillation~\cite{shridhar-etal-2023-distilling}, code generation studies introduce solution plans or intermediate reasoning information as training supervision to improve code generation models~\cite{sun-etal-2024-enhancing-code}. COTTON~\cite{yang2024cotton} trains lightweight models to generate CoTs for code generation, while MapCoder-Lite~\cite{lee-etal-2026-mapcoder} distills knowledge from multi-stage planning-and-coding workflows into smaller models.

In addition to using intermediate information as direct supervision, recent studies further optimize models based on the solution process itself. ReCode~\cite{fan-etal-2026-recode} introduces process-level signals during training, while other approaches~\cite{yang2026thinklike} use structured intermediate objectives to better connect solution formulation with generated code. Another related direction focuses on improving how coders utilize externally provided plans. PADA-Coder~\cite{huang2026padacoder} explicitly trains coders to better follow generated plans.

Existing approaches aim to tie model training to planning information, either by using plans as training signals or by optimizing models to utilize planning-related information. Our work differs by decoupling coder improvement from planning supervision and examining whether planner-agnostic coder-side improvements can generalize across different planners.
\section{Conclusion}
In this paper, we investigate whether planning and coding in LLM-based code generation should be considered as a unified capability or two separable capabilities. Our study shows that these capabilities can exhibit different strengths across models and can be independently utilized to improve code generation. This separation provides new opportunities for constructing code generation systems by combining complementary models and improving individual components without requiring joint optimization of the entire pipeline.

\bibliographystyle{ACM-Reference-Format}
\bibliography{references}

\end{document}